\documentclass[11pt,a4paper]{article}
\usepackage[hyperref]{emnlp2018}
\usepackage{times}
\usepackage{latexsym}
\usepackage{amsmath, amssymb}
\usepackage{graphicx}
\usepackage{bm}
\usepackage{url}
\usepackage{enumitem}

\aclfinalcopy 

\title{Neural Latent Relational Analysis to Capture Lexical Semantic Relations in a Vector Space}

\author{Koki Washio$^{\ast\dag1}$ \and Tsuneaki Kato$^{\ast2}$ \\
		$^\ast$Department of Language and Information Sciences,  The University of Tokyo \\
        $^\dag$RIKEN Center for Advanced Intelligence Project \\
		\tt{\{$^1$kokiwashio@g.ecc, $^2$kato@boz.c\}.u-tokyo.ac.jp}}

\date{}

\begin{document}
\maketitle
\begin{abstract}
Capturing the semantic relations of words in a vector space contributes to many natural language processing tasks.
One promising approach exploits lexico-syntactic patterns as features of word pairs.
In this paper, we propose a novel model of this pattern-based approach, neural latent relational analysis (NLRA). NLRA can generalize co-occurrences of word pairs and lexico-syntactic patterns, and obtain embeddings of the word pairs that do not co-occur.
This overcomes the critical data sparseness problem encountered in previous pattern-based models.
Our experimental results on measuring relational similarity demonstrate that NLRA outperforms the previous pattern-based models.
In addition, when combined with a vector offset model, NLRA achieves a performance comparable to that of the state-of-the-art model that exploits additional semantic relational data.
\end{abstract}

\section{Introduction}
Information on the semantic relations of words is important for many natural language processing tasks, such as recognizing textual entailment, discourse classification, and question answering.
There are two main approaches to obtain the distributed relational representations of word pairs.

One is the vector offset method \cite{mikolov2013distributed, mikolov2013}.
This approach represents word pairs as the vector offsets of their word embeddings.
Another approach exploits lexico-syntactic patterns to obtain word pair representations.
As a pioneer work, \newcite{turney2005} introduced latent relational analysis (LRA), based on the {\it latent relation hypothesis}. It states that word pairs that co-occur in similar lexico-syntactic patterns tend to have similar semantic relations \cite{turney2008latent,turney2010frequency}.
LRA is expected to complement the vector offset model because word embeddings do not contain information on lexico-syntactic patterns that connect word pairs in a corpus \cite{shwartz2016}.

However, LRA cannot obtain the representations of word pairs that do not co-occur in a corpus.
Even with a large corpus, observing a co-occurrence of all semantically related word pairs is nearly impossible because of Zipf's law, which states that most content words rarely occur.
This data sparseness problem is a major bottleneck of pattern-based models such as LRA.

In this paper, we propose neural latent relational analysis (NLRA) to solve that data sparseness problem.
NLRA unsupervisedly learns the embeddings of target word pairs and co-occurring patterns from a corpus.
In addition, it jointly learns the mapping from the word embedding space to the word-pair embedding space.
By this mapping, NLRA can generalize the co-occurrences of word pairs and patterns, and obtain the relational embeddings for arbitrary word pairs even if they do not co-occur in the corpus.

Our experimental results on the task of measuring relational similarity show that NLRA significantly outperforms LRA, and it can also capture semantic relations of word pairs without co-occurrences.
Moreover, we show that combining NLRA and the vector offset model improves the performance and leads to competitive results to those of the state-of-the-art method that exploits additional semantic relational data.

\begin{figure*} 
\centering
\includegraphics[width=11cm]{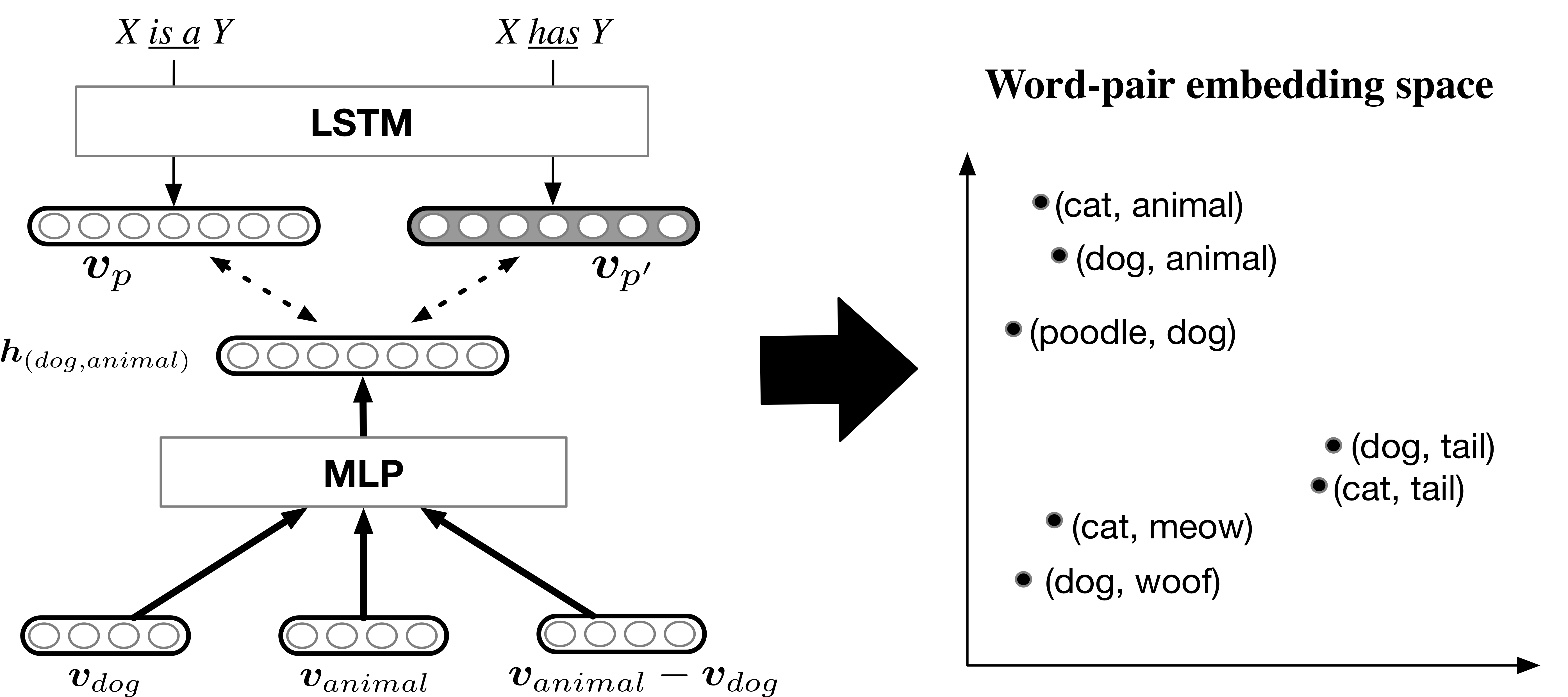}
\caption{An illustration of NLRA}
\label{fig:nlra}
\end{figure*}

\section{Background}

\subsection{Vector Offset Model} \label{sec:vector offset}
The vector offset model \cite{mikolov2013distributed, mikolov2013,levy2014} obtains word embeddings from a corpus and represents each word pair $(a, b)$ as the vector offset of their embedding as follows:
\begin{equation}
	{\bm v}_{(a, b)} = {\bm v}_{b} - {\bm v}_{a}
\end{equation}
where ${\bm v}_{a}$ and ${\bm v}_{b}$ are the word embeddings of $a$ and $b$ respectively.

This method regards relational information as the change in multiple topicality dimensions from one word to the other in the word embedding space \cite{zhila2013}.
Meanwhile, it does not contain the information of lexico-syntactic patterns that were shown to capture complementary information with word embeddings in previous studies on the lexical semantic relation detection \cite{levy2015, shwartz2016}.

\subsection{Latent Relational Analysis} \label{sec:lra}
LRA takes a set of word pairs as input and generates the distributed representations of those word pairs based on their co-occurring patterns.

Given target word pairs $W = \{(a_1,b_1), \ldots, (a_n,b_n)\}$, LRA constructs a list of lexico-syntactic patterns that connect those pairs, such as {\it is a} or {\it in the}, from the corpus for each word pair.
Then, those patterns are generalized by replacing any or all or none of the intervening words with wildcards.
As a feature selection, the generalized patterns generated from many word pairs are used as features.
We define the set of these target feature patterns as $C = \{p_1,  \ldots, p_{m}\}$.
Then, the $2n \times 2m$ matrix $M$ is constructed.
The rows of $M$ correspond to pairs $(a_i, b_i)$ and reversed pairs $(b_i, a_i)$.
The columns of $M$ correspond to patterns $X p_i Y$ and swapped patterns $Y p_i X$, where $X$ and $Y$ are the slots for the words of the word pairs.
The value of $M_{ij}$ represents the strength of the association between the corresponding word pair and pattern, which is calculated using weighting methods such as positive pointwise mutual information (PPMI).
After these processes, the singular value decomposition (SVD) is applied to $M$, and the vector ${\bm v}_{(a, b)}$ is assigned to each word pair $(a, b)$.

Although pattern-based approaches such as LRA have achieved promising results in some semantic relational tasks \cite{turney2008uniform, turney2008latent}, they have a crucial problem that a co-occurrence of all semantically related word pairs cannot be observed because of Zipf's law, which states that the frequency distribution of words has a long tail. In other words, most words occur very rarely \citep{hanks:2009}.
For the word pairs without co-occurrences, LRA cannot obtain their vector representations.

\section{Neural Latent Relational Analysis} \label{sec:nlra}

We introduce NLRA, based on the latent relation hypothesis.
NLRA represents the target word pairs and lexico-syntactic patterns as embeddings.
Similar to the skip-gram model \cite{mikolov2013distributed}, 
NLRA updates those representations unsupervisedly, such that the inner products of the word pairs and patterns in which they co-occur in a corpus have high values.
Through this learning, the word pairs that co-occur in similar patterns have similar embeddings.
Moreover, NLRA can generalize the co-occurrences of the word pairs and patterns by constructing the embeddings of the word pairs from their word embeddings, thus solving the data sparseness problem of word co-occurrences.
Therefore, NLRA can provide representations that capture the information of lexico-syntactic patterns even for the word pairs that do not co-occur in a sentence.

Figure \ref{fig:nlra} is an illustration of our model. 
NLRA encodes a word pair $(a, b)$ into a dense vector as follows:
\begin{equation}
	\bm{h}_{(a,b)} = MLP([\bm{v}_{a}; \bm{v}_{b}; \bm{v}_{b} - \bm{v}_{a}])
\end{equation}
where $[\bm{v}_{a}; \bm{v}_{b}; \bm{v}_{b} - \bm{v}_{a}]$ is the concatenation of the word embeddings of $a$ and $b$ and their vector offsets; $MLP$ is a multilayer perceptron with nonlinear activation functions.

A pattern $p$ is a sequence of the words $w_1, \ldots, w_{k}$.
The sequence of the corresponding word embeddings ${\bm w}_1, \ldots, {\bm w}_{k}$ are encoded using long short-term memory (LSTM) \cite{hochreiter1997}.
Then, the final output vector ${\bm v}_{p}$ is used as the pattern embedding.

For unsupervised learning, we use the negative sampling objective \cite{mikolov2013distributed}.
Given a set of observed triples $(a, b, p) \in D$, where $a$ and $b$ are words such that $(a, b) \in W$, or $(b, a) \in W$ and $p$ is a co-occurring pattern from a corpus, the objective is as follows:
\begin{eqnarray}
	L &=& \sum_{(a, b, p) \in D} \log \sigma(\bm{v}_{p} \cdot \bm{h}_{(a, b)})  \nonumber \\
    &+& \sum_{(a, b, p') \in D'} \log \sigma(- \bm{v}_{p'} \cdot \bm{h}_{(a, b)})
\end{eqnarray}
\normalsize
where $D'$ is a set of randomly generated negative samples and $\sigma$ is the sigmoid function.
We sampled 10 negative patterns for each word pair.
This objective is maximized using the stochastic gradient descent.

After unsupervised learning, we can obtain word pair representations $\bm{v}_{(a, b)}$ as follows:
\begin{equation}
	\bm{v}_{(a, b)} = [\bm{h}_{(a, b)}; \bm{h}_{(b, a)}]
\end{equation}

\section{Evaluation}
\subsection{Dataset}
In our evaluation, we used the SemEval-2012 Task 2 dataset \cite{semeval-2012} for the task of measuring relational similarity.
This dataset contains a collection of 79 fine-grained semantic relations.
For each relation, there are a few prototypical word pairs and a set of several dozen target word pairs.
The task is to rank the target pairs based on the extent to which they exhibit the relation.
In our experiment, we calculated the score of a target word pair with the average cosine similarity between it and each prototypical word pair.
The models are evaluated in terms of the MaxDiff accuracy and Spearman's correlation.
Following previous works \cite{rink2012,zhila2013}, we used the test set that includes 69 semantic relations to evaluate the performance.

\begin{table*}[t]
\centering
\small
\scalebox{0.9}{
\begin{tabular}{l|cccc||cccc|}
\cline{2-9}
                                      & \multicolumn{4}{c||}{Accuracy}                           & \multicolumn{4}{c|}{Correlation}                         \\ \hline
\multicolumn{1}{|c|}{Relation}           & VecOff        & LRA   & NLRA           & NLRA+VecOff    & VecOff         & LRA   & NLRA           & NLRA+VecOff    \\ \hline
\multicolumn{1}{|l|}{Class-Inclusion} & 0.543         & 0.485 & 0.533          & \textbf{0.56}  & 0.487          & 0.427 & \textbf{0.622} & 0.611          \\
\multicolumn{1}{|l|}{Part-Whole}      & 0.45          & 0.427 & 0.465          & \textbf{0.488} & 0.304          & 0.282 & 0.38           & \textbf{0.395} \\
\multicolumn{1}{|l|}{Similar}         & 0.414         & 0.346 & 0.412          & \textbf{0.436} & 0.267          & 0.123 & 0.271          & \textbf{0.315} \\
\multicolumn{1}{|l|}{Contrast}        & 0.343         & 0.349 & \textbf{0.377} & 0.374          & 0.108          & 0.065 & 0.092          & \textbf{0.124} \\
\multicolumn{1}{|l|}{Attribute}       & 0.462         & 0.414 & 0.447          & \textbf{0.486} & 0.406          & 0.299 & 0.367          & \textbf{0.456} \\
\multicolumn{1}{|l|}{Non-Attribute}   & \textbf{0.39} & 0.366 & 0.369          & 0.381          & \textbf{0.217} & 0.16  & 0.125          & 0.174          \\
\multicolumn{1}{|l|}{Case Relations}  & 0.468         & 0.438 & 0.536          & \textbf{0.558} & 0.391          & 0.291 & \textbf{0.553} & 0.544          \\
\multicolumn{1}{|l|}{Cause Purpose}   & 0.444         & 0.471 & 0.448          & \textbf{0.485} & 0.345          & 0.387 & 0.397          & \textbf{0.454} \\
\multicolumn{1}{|l|}{Space-Time}      & 0.5           & 0.428 & 0.516          & \textbf{0.525} & 0.424          & 0.31  & 0.489          & \textbf{0.493} \\
\multicolumn{1}{|l|}{Reference}       & 0.441         & 0.447 & 0.449          & \textbf{0.465} & 0.297          & 0.346 & \textbf{0.404}          & 0.378 \\ \hline
\multicolumn{1}{|l|}{Average}         & 0.443         & 0.415 & 0.453          & \textbf{0.475} & 0.321          & 0.246 & 0.36           & \textbf{0.391} \\ \hline
\end{tabular}
}
\caption{Average MaxDiff accuracy and Spearman's correlation of each major relation group.}
\label{table:result}
\end{table*}

\subsection{Baselines}

\begin{description}[style=unboxed,leftmargin=0cm]
\item[VecOff.]
We used the 300-dimensional pre-trained GloVe \cite{pennington2014}\footnote{https://nlp.stanford.edu/projects/glove/} and represented word pairs as described in Section \ref{sec:vector offset}.

\item[LRA.]
We implemented LRA as described in Section \ref{sec:lra}.
We set $W$ as the lemmatized word pairs of the dataset.
We used the English Wikipedia as a corpus.
For each word pair, we searched for patterns of from one to three words. When searching for patterns, the left word and right word adjacent to the patterns were lemmatized to ignore their inflections.
Following \cite{turney2008latent}, we selected $C$ as the top $20|W|$ generalized patterns.
Then, $M$ was constructed using PPMI weighting, and its dimensionality was reduced to 300 using SVD.
\end{description}

\subsection{Our methods}
\begin{description}[style=unboxed,leftmargin=0cm]
\item[NLRA.]
For each word pair in the dataset, co-occurring patterns were extracted from the same corpus in the same manner as with LRA, resulting in $D$.
For word embeddings, we used the same pre-trained GloVe as VecOff.
These embeddings were updated during the training.
For $MLP$, we used three affine transformations followed by the batch normalization \cite{ioffe:2015} and tanh activation.
The size of each hidden layer of the MLP was 300.
To encode the patterns, we used LSTM with the 300-dimensional hidden state.
The objective was optimized by AdaGrad \cite{duchi2011} (whose learning rate was 0.01).
We trained the model for 50 epochs.

\item[NLRA+VecOff.]
This method combines NLRA and VecOff by averaging their score for a target word pair.

\end{description}

\subsection{Result and Analysis}

\begin{table}
\centering
\scalebox{0.85}{
\small
\begin{tabular}{|l|ccc|}
\hline
\multicolumn{1}{|c|}{Pair} & Human & LRA                     & NLRA  \\ \hline
laugh:happiness           & 50    & 0.217                   & 0.578 \\
nod:agreement             & 46    & 0.245                   & 0.347 \\
tears:sadness             & 44    & 0.381                   & 0.483 \\ \hline
\multicolumn{1}{|c|}{$\cdots$} & \multicolumn{3}{c|}{$\cdots$}                \\ \hline
scream:terror             & 26    & 0.396                   & 0.417 \\
handshake:cordiality      & 24    & \textbf{0 (no pattern)} & 0.34  \\
lie:dishonesty            & 16    & 0.206                   & 0.394 \\ \hline
\multicolumn{1}{|c|}{$\cdots$} & \multicolumn{3}{c|}{$\cdots$}                \\ \hline
discourse:relationship    & -60   & 0.331                   & 0.275 \\
friendliness:wink         & -68   & \textbf{0 (no pattern)} & 0.26  \\ \hline
\end{tabular}
}
\caption{The scores assigned by humans, LRA, and NLRA for the Reference-Express relation. The pairs are sorted in descending order according to the human score.}
\label{tab:analysis}
\end{table}

Table \ref{table:result} displays the overall result.

\subsubsection*{NLRA vs. LRA}
First, NLRA outperformed LRA in terms of both the average accuracy and correlation.
These differences were statistically significant ($p<0.01$) with the paired t-test.
These results indicate that generalizing patterns with LSTM is better than by using wildcards.
Moreover, NLRA can successfully calculate the relational similarity for the word pairs that do not co-occur in the corpus.
Table \ref{tab:analysis} shows an example of the Reference--Express relation, where the middle-score pair \textit{handshake:cordiality} and the low-score pair \textit{friendliness:wink} have no co-occurring pattern.
In these cases, LRA could not obtain the representations of those word pairs nor correctly assign the score.
By contrast, NLRA could accomplish both because it could generalize the co-occurrences of word pairs and patterns.

\subsubsection*{NLRA+VecOff vs. Other Models}
Second, NLRA+VecOff outperformed the other models.
These differences were statistically significant (the correlation difference between NLRA+Vecoff and NLRA: $p<0.05$; the other differences: $p<0.01$).
These results indicate that lexico-syntactic patterns and the vector offset of word embeddings capture complementary information for measuring relational similarity.
This is inconsistent with the findings of \newcite{zhila2013}.
That work combined heterogeneous models, such as the vector offset model, pattern-based model, etc., and stated that the pattern-based model was less significant than the vector offset model, based on their ablation study.
We believe that this was because their pattern-based model did not generalize patterns with wildcards nor select useful features.
Their pattern-based model seemed to suffer from sparse feature space.
In our experiment, 
NLRA helped VecOff, for example, for the Part-Whole relation, Cause Purpose relation, and Space-Time relation, where there seemed to be prototypical patterns indicating those relations.
Meanwhile, VecOff helped NLRA for the Attribute relation, where the relational patterns seemed to be diverse.
These results showed that the combined model is robust.

\subsection{Comparison to other systems}
\begin{table}[t]
\centering
\scalebox{0.85}{
\small
\begin{tabular}{lcc}
\hline
\multicolumn{1}{c}{Model} & \multicolumn{1}{l}{Accuracy} & \multicolumn{1}{l}{Correlation} \\ \hline
\newcite{rink2012}  & 0.394                        & 0.229                           \\
\newcite{mikolov2013}     & 0.418                        & 0.275                           \\
\newcite{levy2014}   & 0.452                        & --                               \\
\newcite{zhila2013}       & 0.452                        & 0.353                           \\
\newcite{iacobacci2015}   & 0.459                        & 0.358                           \\
\newcite{turney2013}             & 0.472                        & \textbf{0.408}                  \\ \hline
VecOff                    & 0.443                        & 0.321                           \\
LRA                       & 0.415                        & 0.264                           \\
NLRA                      & 0.453                        & 0.36                            \\ 
NLRA+VecOff               & \textbf{0.475}               & 0.391	\\ \hline
\end{tabular}
}
\caption{Published results of other models on the SemEval2012 Task 2 dataset.}
\label{tab:comparison}
\end{table}

We compared the results of our models to other published results.
Table \ref{tab:comparison} displays those results.
\newcite{rink2012} is the pattern-based model with naive Bayes.
\newcite{mikolov2013}, \newcite{levy2014}, and \newcite{iacobacci2015} are the vector offset models.
\newcite{zhila2013} is the model composed of various features.
\newcite{turney2013} extracts the statistical features of two word pairs from a word-context co-occurrence matrix and trains the classifier with additional semantic relational data to assign a relational similarity for two word pairs.

NLRA+VecOff achieved a competitive performance to the state-of-the-art method of \newcite{turney2013}.
Note that our method learns unsupervisedly and does not exploit additional resources, and the method of \newcite{turney2013} cannot obtain the distributed representation of word pairs.

A work similar to ours, \newcite{bollegala2015}, represented lexico-syntactic patterns as the vector offset of co-occurring word pairs and updated the vector offsets of word pairs such that word pairs that co-occur in similar patterns have similar offsets. They evaluated their model on all 79 semantic relations of the dataset and achieved 0.449 accuracy.
In their setting, NLRA+VecOff achieved 0.47 accuracy, outperforming their model.

\section{Related Work}
\subsection{Word Pairs and Co-occurring Patterns}
\citet{hearst1992} detected the hypernymy relation of word pairs from a corpus using several handcrafted lexico-syntactic patterns.
\citet{turney2005corpus} used 64 handcrafted lexico-syntactic patterns as features of word pairs to represent word pairs as vectors.
To obtain word-pair embeddings, \citet{turney2005} extended the method of \citet{turney2005corpus} as LRA.
Our work is a neural extension of LRA.

\citet{washio2018filling} proposed the method similar to ours in lexical semantic relation detection.
Their neural method modeled the co-occurrences of word pairs and dependency paths connecting two words to alleviate the data sparseness problem of pattern-based lexical semantic relation detection.
While they assigned randomly initialized embeddings to each dependency path, our work encodes co-occurring patterns with LSTM for better generalization.
\citet{jameel2018unsuoervised} embedded word pairs with the context words occurring around word pairs instead of lexico-syntactic patterns.
Their method cannot obtain embeddings of word pairs that do not co-occur in a corpus because they directly assigned embeddings to word pairs.
By contrast, NLRA can obtain embeddings for those word pairs.

In another research area, relation extraction, several works have explored an idea similar to the latent relation hypothesis \cite{riedel2013relation, toutanova2015representing, verga2017generalizing}. 
They factorized a matrix of entity pairs and co-occurring patterns, while they focused on named entity pairs instead of word pairs and did not consider co-occurrence frequencies.

\subsection{Relation to Knowledge Graph Embedding}
Knowledge graph embedding (KGE) embeds entities and relations in knowledge graph (KG), where entities and relations corresponds to nodes and edges respectively \cite{nickel2011three,bordes2013translating,socher2013reasoning, wang2014knowledge, lin2015learning, yangl2015embedding,nickel2016holographic,trouillon2016complex, liu2017analogical, wang2017knowledge, ishihara2018neural}.
By considering words and lexico-syntactic patterns as nodes and edges, respectively, a corpus can be viewed as a graph, i.e., corpus graph (CG).
Thus, NLRA can be regarded as corpus graph embedding (CGE) models based on the latent relation hypothesis.

Although KGE models can be easily applied to CG,
several differences exist between KG and CG.
First, the nodes and edges of CG are (sequences of) linguistic expressions, such as tokens, lemmas, phrases, etc.
Thus, the nodes and edges of CG might exhibit compositionality and ambiguity, while KG does not have those properties.
Second, the edges of CG have weights based on co-occurrence frequencies unlike the edges of KG.
Finally, CG might have a large number of edges types while the number of KG edges is at most several thousands.
An interesting research direction is exploring models suitable for CGE to capture the property of linguistic expressions and their relations in the embedding space.

\section{Conclusion}
We presented NLRA, which learns the distributed representation of word pairs capturing semantic relational information through co-occurring patterns encoded by LSTM.
This model jointly learns the mapping from the word embedding space into the word-pair embedding space to generalize co-occurrences of word pairs and patterns.
Our experiment on measuring relational similarity demonstrated that NLRA outperforms LRA and can successfully solve the data sparseness problem of word co-occurrences, which is a major bottleneck in pattern-based approaches.
Moreover, combining the vector offset model and NLRA yielded competitive performance to the state-of-the-art method, though our method relied only on unsupervised learning.
This combined model exploits the complementary information of lexico-syntactic patterns and word embeddings.

In our future work, we will apply word-pair embeddings from NLRA to various downstream tasks related to lexical relational information.

\section*{Acknowledgments}
This work was supported by JSPS KAKENHI Grant numbers JP17H01831.
We thank Satoshi Sekine and Kentaro Inui for helpful discussion.

\bibliography{emnlp2018}
\bibliographystyle{acl_natbib_nourl}

\end{document}